\documentclass[10pt,twocolumn,letterpaper]{article}

\usepackage[pagenumbers]{cvpr} %

\definecolor{cvprblue}{rgb}{0.21,0.49,0.74}
\usepackage[pagebackref,breaklinks,colorlinks,allcolors=cvprblue]{hyperref}

\usepackage{caption}
\usepackage{graphics} %
\usepackage{graphicx}
\usepackage{mathptmx} %
\usepackage{amsmath} %
\usepackage{amssymb}  %
\usepackage{algorithm}
\usepackage{algpseudocode}
\usepackage{booktabs}
\usepackage{xcolor}

\def\paperID{13044} %
\def\confName{CVPR}
\def\confYear{2025}

\title{Text-to-3D Gaussian Splatting with Physics-Grounded Motion Generation}

\author{Wenqing Wang, Yun Fu\\
Northeastern University, USA\\
360 Huntington Ave, Boston, MA 02115\\
{\tt\small wang.wenqin@northeastern.edu, yunfu@ece.northeastern.edu}
}

\begin{document}

\twocolumn[{%
\renewcommand\twocolumn[1][]{#1}%
\maketitle
\begin{center}
    \centering
    \captionsetup{belowskip=-10pt} 
    \captionsetup{type=figure}
    \includegraphics[width=\textwidth, trim=0 0 0 0, clip]{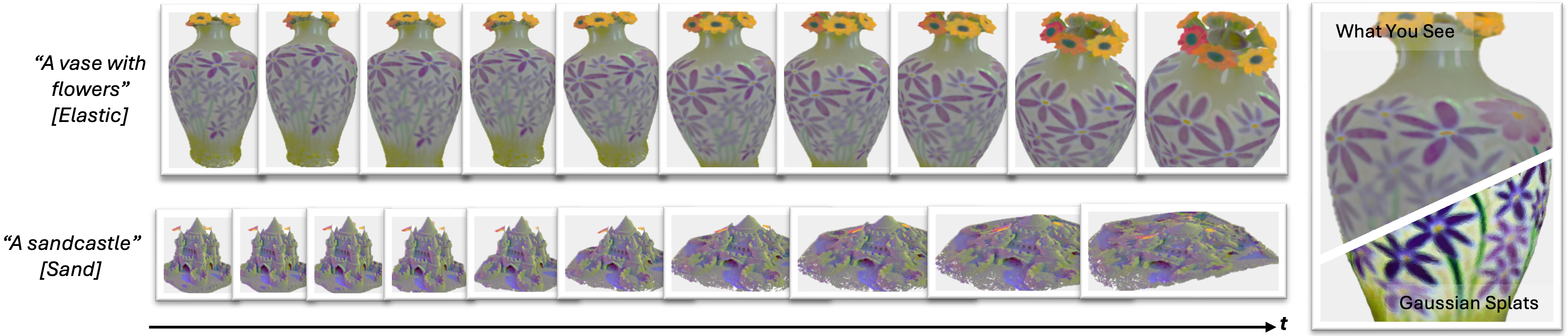}
    \captionof{figure}{Our framework is a text-to-3D physics-grounded motion-rendering pipeline with high-quality visual appearances and realistic motion.}
\label{fig:teaser}
\end{center}
}]

\begin{abstract}
Text-to-3D generation is a valuable technology in virtual reality and digital content creation. While recent works have pushed the boundaries of text-to-3D generation, producing high-fidelity 3D objects with inefficient prompts and simulating their physics-grounded motion accurately still remain unsolved challenges. To address these challenges, we present an innovative framework that utilizes the Large Language Model (LLM)-refined prompts and diffusion priors-guided Gaussian Splatting (GS) for generating 3D models with accurate appearances and geometric structures. We also incorporate a continuum mechanics-based deformation map and color regularization to synthesize vivid physics-grounded motion for the generated 3D Gaussians, adhering to the conservation of mass and momentum. By integrating text-to-3D generation with physics-grounded motion synthesis, our framework renders photo-realistic 3D objects that exhibit physics-aware motion, accurately reflecting the behaviors of the objects under various forces and constraints across different materials. Extensive experiments demonstrate that our approach achieves high-quality 3D generations with realistic physics-grounded motion.
\end{abstract}

\section{Introduction}
\label{sec:intro}

Text-to-3D modeling has demonstrated remarkable achievements in creating highly realistic 3D representations of objects. Recently, several works have made great progress in generating delicate 3D objects using text-to-image priors \cite{hong20243dtopialargetextto3dgeneration, Ding_2024_CVPR, Wu_2024_CVPR, yan2024dreamdissectorlearningdisentangledtextto3d}. Additionally, other works have strides in producing the motion of the given 3D objects \cite{Kapon_2024_CVPR, 10550606, 10550763, Guo_2024_CVPR, Liang_2024_CVPR1}. Despite these advancements, current methods face challenges in synthesizing realistic 3D objects from inefficient text prompts and accurately simulating their physics-grounded motion.

3D Gaussian Splatting \cite{kerbl20233dgaussiansplattingrealtime} has become a prominent technique in the domain of neural rendering, due to its remarkable ability to render delicate details, point-based representation, and rapid rendering speed. Several works have leveraged 3D GS to generate photo-realistic 3D models from the text prompts \cite{kerbl20233dgaussiansplattingrealtime, zhou2024gala3dtextto3dcomplexscene, Chen_2024_CVPR, Yi_2024_CVPR, Liu_2024_CVPR, Liang_2024_CVPR2}. A notable work GSGEN \cite{chen2024textto3dusinggaussiansplatting} integrates 3D GS with diffusion priors to produce 3D objects with highly realistic structures and visual fidelity. Other works adopt 3D Gaussian representations to model dynamic motion \cite{zhu2024motiongsexploringexplicitmotion, guo2024motionaware3dgaussiansplatting, lee2024crimgscontinuousrigidmotionaware, Huang_2024_CVPR}. Xie \textit{et al.} introduces a remarkable framework PhysGaussian \cite{xie2024physgaussianphysicsintegrated3dgaussians} that utilizes the physics models that describe the materials' behaviors to guide the 3D GS to simulate the object motion. These works have laid a robust foundation for the integration of text-to-3D generation and 3D-to-motion simulations.

However, current works have not fully explored techniques for producing high-quality 3D models with realistic, physics-grounded motion from text prompts. In addition, existing text-to-3D frameworks are often guided by text-to-2D image generation models, which have limited text-understanding ability. This limitation can lead to unsatisfied 3D generations when given poorly written text prompts. To overcome these challenges, we introduce a new framework that enables text-to-3D generation of physics-grounded motion with the aid of LLM-based prompt refinement. To achieve this, we utilize an LLM to refine the input text prompts. Then, we adopt 3D Gaussians as our 3D object representations and use the 3D (shape) diffusion prior and 2D (image) diffusion prior to guide the 3D GS to create photorealistic 3D models with reasonable geometric shapes and realistic appearances. Furthermore, we simulate physics-grounded motion on the generated 3D Gaussians by using a continuum mechanics-based deformation map to deform the Gaussian kernels. Additionally, we introduce a color regularization technique to ensure that the rendered objects maintain accurate and consistent colors. As a result, our framework generates high-quality 3D objects that exhibit physics-grounded motion. In conclusion, our main contributions include:
\begin{itemize}
  \item We present an innovative framework for synthesizing high-quality 3D objects with realistic, physics-based motion derived from text prompts.

  \item We leverage an LLM to refine text prompts and diffusion priors to guide the generation of geometrically accurate and visually appealing 3D models.
  
  \item We utilize a continuum mechanics-based deformation map combined with a color regularization technique to produce realistic 3D object motion with accurate colors.
\end{itemize}

\section{Related Work}
\label{sec:related_work}

\subsection{Neural Rendering}
Recent breakthroughs in neural rendering have significantly impacted novel view synthesis. Rendering with radiance fields gained considerable interest due to their remarkable ability to synthesize novel views and their significant promise for advancing 3D generative tasks. Building on this foundation, Neural Radiance Fields (NeRF) \cite{mildenhall2020nerfrepresentingscenesneural} revolutionizes volumetric rendering by leveraging neural networks to encode 3D scenes, achieving impressive rendering results. Subsequent works have emerged to improve NeRF in tasks such as 3D scene reconstruction \cite{rs15143585, Chen_2023_ICCV, 10704527, NEURIPS2021_7d62a275}, in-the-wild scene handling \cite{Martin-Brualla_2021_CVPR, Chen_2022_CVPR, 10.1145/3528233.3530718}, training speed optimization \cite{Deng_2022_CVPR, Wang_2023_CVPR, Li_2023_ICCV}, and rendering quality improvement \cite{10.1145/3503161.3547808, 10742507, Huang_2023_CVPR}. However, NeRF poses a computational challenge because of the extensive sampling required along each ray, leading to slow rendering speed and high memory consumption. To overcome these challenges, a point-based rendering method 3D Gaussian Splatting \cite{kerbl20233dgaussiansplattingrealtime} is introduced to represent scenes with 3D Gaussians and render with a fast rasterization method. This enables it to achieve both rapid rendering and high-quality generation. In our proposed framework, we leverage 3D Gaussians to represent the 3D objects and utilize its point-based nature to generate consistent shapes and realistic motion.

\subsection{Text-to-3D Generation}
As a groundbreaking approach in generative AI, text-to-3D generation enables synthesizing 3D models directly from the input text prompts. With the recent advancement in diffusion models, a wave of advancements in text-to-3D generation utilize diffusion priors to guide the 3D generation to be aligned with the text prompt description \cite{Xu_2023_CVPR, hong20243dtopialargetextto3dgeneration, Ding_2024_CVPR, Tang_2023_ICCV, li2023sweetdreameraligninggeometricpriors, raj2023dreambooth3dsubjectdriventextto3dgeneration, chen2024textto3dusinggaussiansplatting}. DreamBooth3D \cite{raj2023dreambooth3dsubjectdriventextto3dgeneration} proposes an efficient optimization strategy that leverages the 3D consistency of NeRF with the 2D diffusion prior. DreamFusion \cite{poole2022dreamfusiontextto3dusing2d} utilizes a score distillation sampling loss to align 2D diffusion prior with the generated images during optimization. Magic3D \cite{lin2023magic3dhighresolutiontextto3dcontent} employs a coarse-to-fine optimization process, incorporating diffusion priors to accelerate NeRF's optimization and improve the quality of generated results. Building on these foundational works, our approach also utilizes diffusion priors to guide 3D generation, focusing on producing high-quality 3D objects with realistic appearances and well-defined shapes.

\subsection{The Material Point Method}
The Material Point Method (MPM) is a computational framework designed to simulate material behaviors by integrating both particle and grid-based approaches \cite{Sulsky_1994}. The MPM represents materials as a set of particles that are mapped onto a grid to compute and simulate the material deformations, which enables the simulation of diverse material properties \cite{ZHENG2022104771, LEI2022105009, HARRIS2021102602, ZHENG20242341}. Owing to these advantages, we utilize the MPM to generate the physics-grounded motion for our 3D objects.

\section{Method}
\label{sec:method}

\begin{figure*}[htbp]
  \centering
  \includegraphics[width=\textwidth, trim=0 0 0 0, clip]{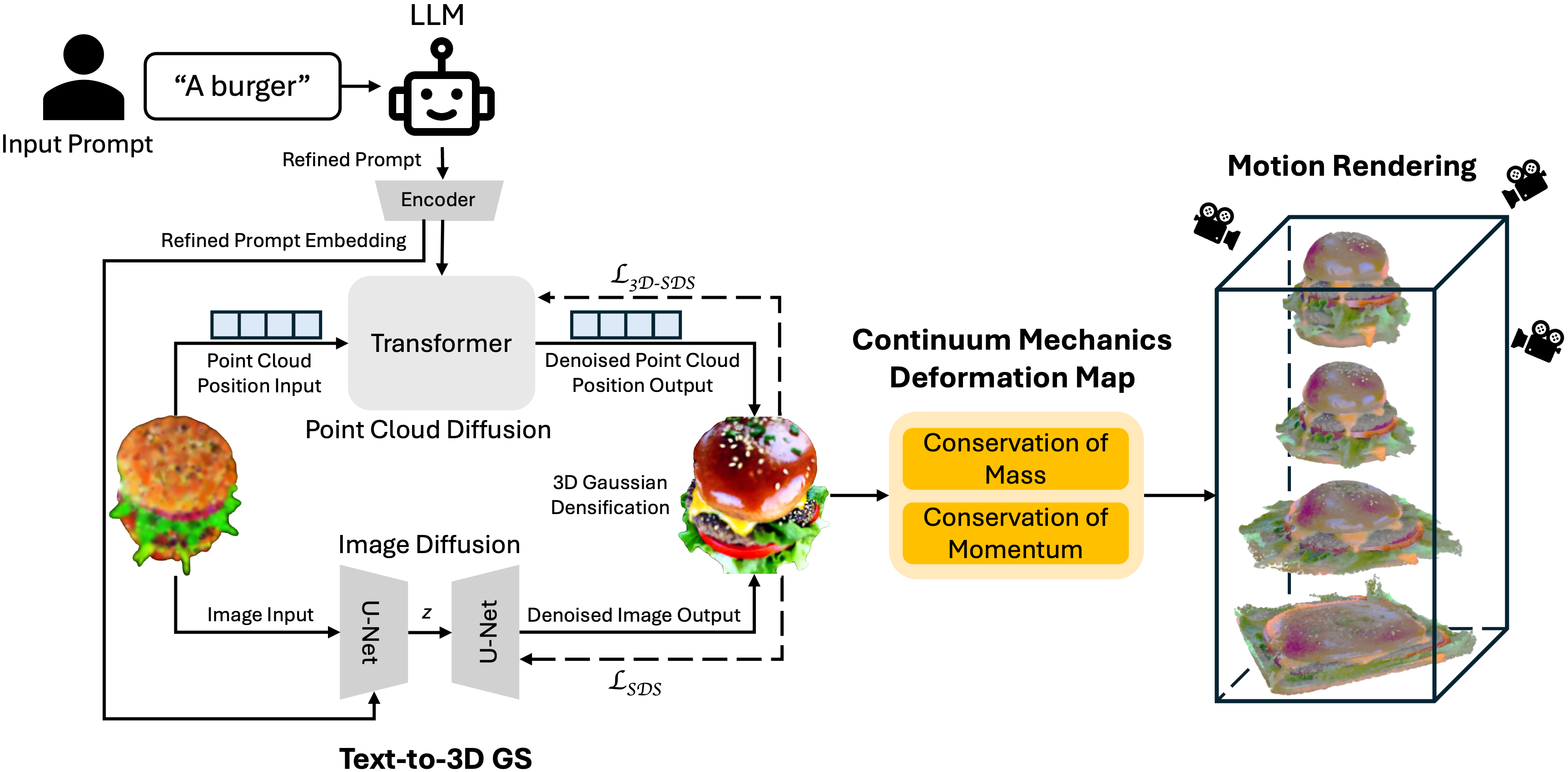}
  \captionsetup{belowskip=-10pt} 
  \caption{\textbf{Pipeline overview.} Our framework first leverages an LLM to refine the text prompt. Next, it employs a 3D geometry diffusion prior and a 2D image diffusion prior for guiding the 3D GS process, producing the high-quality 3D object. Finally, a deformation map based on continuum mechanics is applied to synthesize the physics-grounded motion of the 3D object.}
  \label{Pipeline}
\end{figure*}

We introduce a framework for synthesizing 3D models with physics-grounded motion using LLM-refined prompts, diffusion priors, and a deformation map (Figure~\ref{Pipeline}). We initially employ an LLM to refine the prompt into a more explicit, detailed, and logically coherent form. For efficient and high-quality 3D model generation, we utilize 3D Gaussian splatting as our object representation. To address challenges such as the Janus problem and to improve the accuracy of the generated 3D object's shape and appearance, we incorporate guidance from both a 3D shape diffusion prior and a 2D image diffusion prior. Subsequently, a deformation map grounded in continuum mechanics is applied to the 3D Gaussian kernels, enabling realistic motion rendering that adheres to the principles of mass and momentum conservation. This section presents an in-depth explanation of the proposed framework.

\subsection{3D Gaussian Splatting}

3D Gaussian Splatting achieves high-quality scene reconstruction with fast training and rendering speeds \cite{kerbl20233dgaussiansplattingrealtime}. As a point-based rendering approach, it represents a scene using 3D Gaussians, defined by their position (mean) $x_i$, covariance matrix $\sigma_i$, opacity $\alpha_i$, and spherical harmonic coefficients $c_i$ as $G(x) = e^{-\frac{1}{2} (x)^T \Sigma^{-1} (x)}$. To render a scene, GS first projects the 3D Gaussians into 2D space. To achieve fast rendering, GS employs a tile-based rasterization strategy, which sorts the projected 2D Gaussians based on their depth in view space. Each screen tile is processed by a thread block that loads the Gaussians into shared memory and computes the final pixel colors via alpha-blending:
\begin{equation}
C = \sum_{i \in N} c_i \alpha_i \prod_{j = 1}^{i-1} (1 - \alpha_j),
\end{equation}
where $\alpha$ indicates the opacity, $c_i$ represents the color of each point $i$, and $N$ denotes the total number of tile Gaussians. To produce high-quality radiance field representation, GS conducts optimization using $L_1$ and $D-SSIM$ loss: $L_{GS} = (1 - \lambda)L_1 + \lambda L_{D-\text{SSIM}}$, and it adaptively controls the density of the 3D Gaussians through pruning and densifying processes \cite{kerbl20233dgaussiansplattingrealtime}. To leverage its fast rendering speed and high-quality rendering ability, we integrate 3D Gaussian Splatting into our framework to generate 3D Gaussians as our object representation. We further extend the GS kernel to incorporate time-dependency in $x_i$ and $\sigma_i$, enabling physics-grounded motion and demonstrating the potential of 3D GS for generative tasks.

\subsection{LLM-Prompt Refinement}

Text-to-3D generation often produces suboptimal results when the input prompt is vague, overly complex, or involves intricate logical relationships. This limitation arises primarily from the constrained text comprehension capabilities of the guidance models used in the process. Typically, 3D generation models rely on 2D content generation frameworks, including methods like diffusion models \cite{rombach2022highresolutionimagesynthesislatent, nichol2022glidephotorealisticimagegeneration}. These 2D generation models in turn depend on classifier guidance models like CLIP's text encoder \cite{radford2021learningtransferablevisualmodels}. These classifier guidance models lack advanced natural language understanding capabilities and are trained on datasets with simple textual descriptions that do not contain complex logic or detailed relational information. Hence, the visual concepts they encode are limited, restricting text-to-3D models to perform effectively only with simple prompts. Furthermore, when the prompts are too vague or brief, the 2D generation models may not have enough context to provide accurate or detailed guidance images for the 3D generation models.

However, we notice that Large Language Models have showcased exceptional abilities in text comprehension, processing, and refinement, owing to their Transformer-based architecture and mechanisms like self-attention and contextual embeddings \cite{vaswani2023attentionneed, brown2020languagemodelsfewshotlearners, neelakantan2022textcodeembeddingscontrastive}. Therefore, we leverage an LLM, ChatGPT-4, to refine text prompts for improved text-to-3D generation. Following the LLM prompt engineering practices in the community \cite{white2023promptpatterncatalogenhance}, the revision instruction prompts that we give to the LLM are composed of the \textbf{context} and \textbf{task} components.
\begin{figure}[htbp]
  \centering
  \includegraphics[width=\columnwidth]{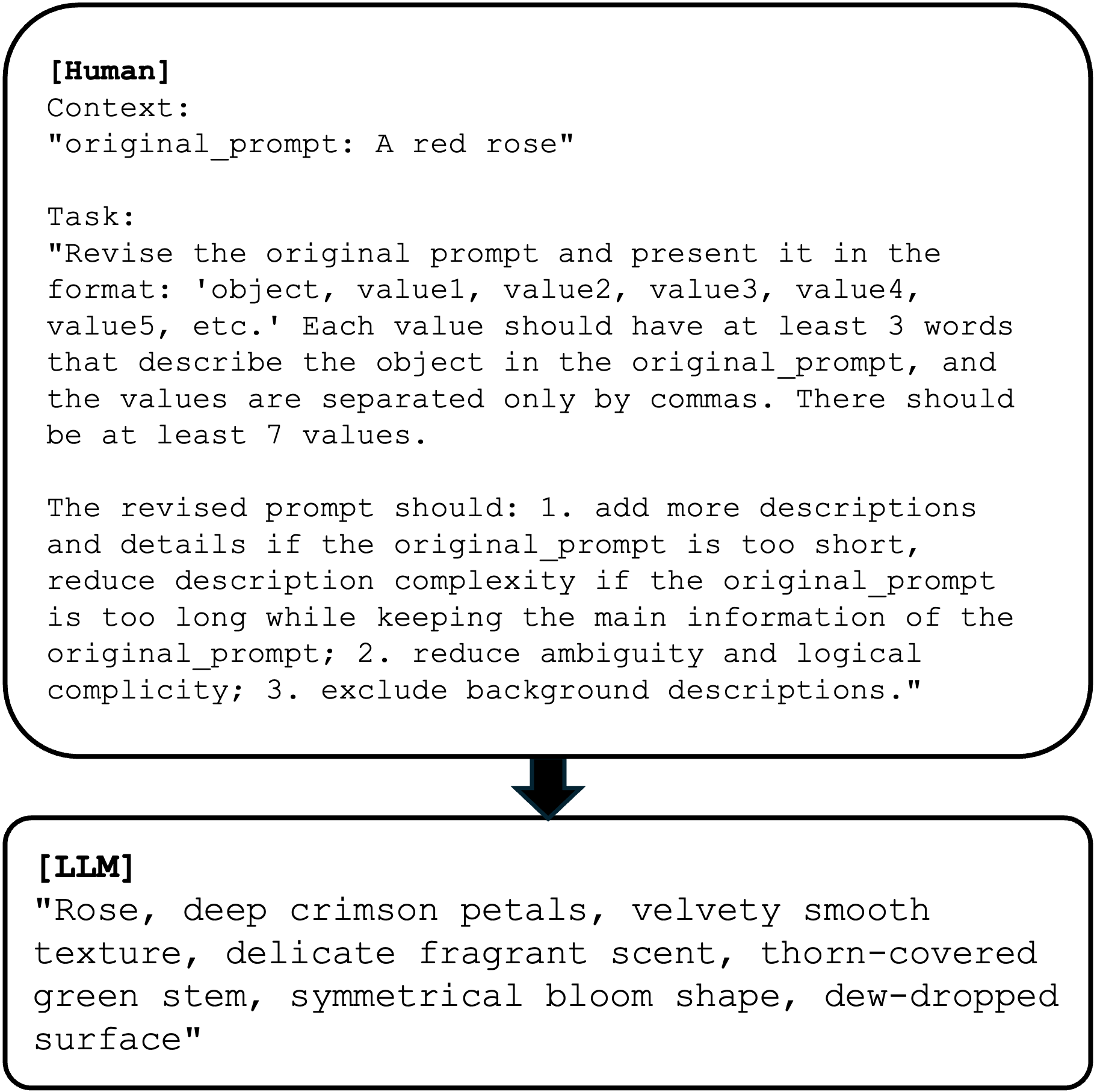}
  \captionsetup{belowskip=-10pt} 
  \caption{LLM-prompt refinement of a vague text prompt.}
  \label{llm_s}
\end{figure}

\begin{figure}[htbp]
  \centering
  \includegraphics[width=\columnwidth]{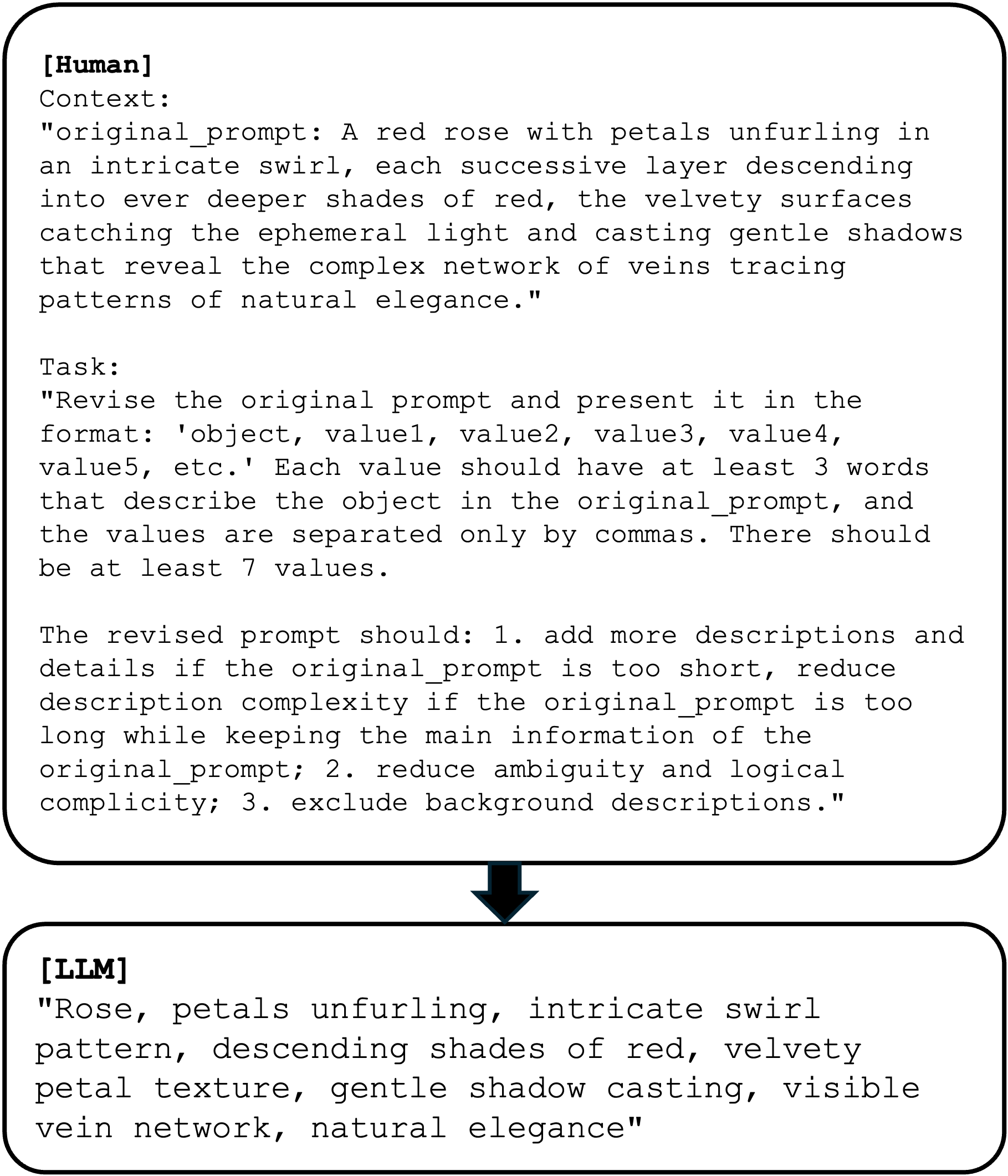}
  \captionsetup{belowskip=-10pt} 
  \caption{LLM-prompt refinement of a complex text prompt.}
  \label{llm_l}
\end{figure}

\noindent \textbf{Context component.} This component contains the original text prompt that needs refinement. As mentioned above, some of the issues regarding the original prompts include vagueness, description complexity, and logical complicity. For example, a prompt might simply describe an object with minimal details, such as ``a red rose" (Figure~\ref{llm_s}). However, this prompt lacks sufficient context and fails to specify details about the rose beyond its color, which can result in generated outputs that are vague or insufficiently detailed. Furthermore, prompts might also be overly lengthy and intricate (Figure~\ref{llm_l}), making it challenging for models to accurately interpret and generate the intended output.

\noindent \textbf{Task component.} The task component provides instructions for refining the original prompt. Specifically, it guides the LLM to revise prompts in a comma-separated format, detailing the characteristics of the target object to improve clarity and interpretability for guidance models. For short and under-described prompts, the revision should include additional details and elaborations to provide richer context. However, if the original prompt is lengthy with complex descriptions, it should simplify the description while preserving the core information. Overall, the revised prompt should address any vagueness and logical inconsistencies. In addition, to emphasize the target object for text-to-3D generation, refined prompts should omit background descriptions.

\subsection{Text-to-3D GS}

To synthesize shape-accurate and visually appealing 3D objects, we leverage 3D Gaussians as our 3D representation using the Gaussian Splatting method. This approach is driven by its point-based structure, capacity to generate high-quality rendering outcomes, and rapid rendering. Building on the work of \cite{chen2024textto3dusinggaussiansplatting}, we iteratively optimize 3D shapes and visual appearances by integrating guidance from diffusion priors. Since 3D Gaussians are created from the point cloud produced by methods such as Structure from Motion, we utilize the 3D diffusion prior from a text-to-3D-point-cloud diffusion framework, \textit{Point-E} \cite{nichol2022pointegenerating3dpoint}. This guides the generation of 3D Gaussians to produce plausible 3D shapes, mitigating the Janus problem—an issue where the model overfits to specific views, resulting in artifacts such as multiple faces or inaccurate geometry. To achieve this, we employ a 3D Score Distillation Sampling (SDS) loss \cite{alldieck2024scoredistillationsamplinglearned} to guide the shape optimization process:
\begin{multline}
    L_{\text{shape}} = \mathbb{E}_{\epsilon_I, t} \left[ w_I(t) \left\| \epsilon_\phi(\hat{I}_t; y, t) - \epsilon_I \right\|^2_2 \right] \\
+ \mathbb{E}_{\epsilon_X, t} \left[ w_X(t) \left\| \epsilon_\psi(x_t; y, t) - \epsilon_X \right\|^2_2 \right] \cdot \lambda_{3D}
\end{multline}
where $x_t$ denotes the noisy Gaussian positions and $\hat{I}$ represents the generated image, $w$ and $\epsilon$ are the weighting function and Gaussian noise.

To improve the visual quality of the generated 3D models, we refine and densify the 3D Gaussians using a 2D diffusion prior derived from a pre-trained 2D image diffusion mode, \textit{Stable-Diffusion} \cite{rombach2022highresolutionimagesynthesislatent}. In this process, the Gaussians gradually improve their visual details and appearances. The loss function of this appearance refinement process is:
\begin{equation}
  L_{\text{appearance}} = \mathbb{E}_{\epsilon, I, t} \left[ w_I(t) \left\| \epsilon_{\phi}(\hat{I}_t; y, t) - \epsilon_I \right\|^2_2 \right] \cdot \lambda_{SDS}
\end{equation}
where $\hat{I}$ is the generated image and $\lambda_{SDS}$ denotes the SDS loss weight. By iteratively optimizing the 3D Gaussians using both the 3D shape diffusion prior and the 2D image diffusion prior, this method enables the synthesis of 3D models that are consistent in shape and visually compelling.

\subsection{Physics-Grounded Deformation Map}

To generate physics-grounded motion in 3D Gaussians, we incorporate continuum mechanics into our framework. Inspired by the work of \cite{xie2024physgaussianphysicsintegrated3dgaussians}, we employ a deformation map $\phi(X, t)$ to describe the motion of a particle's position $x_i$ at the time $t$. Local transformations, such as rotation and stretch at any position, are defined using the deformation map gradient as $F(X, t) = \nabla_X\phi(X, t)$. This can be decomposed as $F = F^EF^P$, where $F^E$ is the elastic part and $F^P$ is the plastic part. To be grounded in continuum mechanics, the updates of the deformation map $\phi$ conform to the principles of mass and momentum conservation\cite{Sulsky_1994}.

Under the mass conservation principle, the material mass should remain constant over time, regardless of how the region deforms:
\begin{equation}
    \int_{R^t_\epsilon} \rho(x, t) \equiv \int_{R^0_\epsilon} \rho(\phi^{-1}(x, t), 0),
\end{equation}
where $\rho(x, t)$ denotes the the material density field and $R^t_\epsilon = \phi(R^0_\epsilon, t)$ is the region within the undeformed material space.

According to the conservation of momentum principle, the momentum of any material region should remain unchanged before and after the deformation:
\begin{equation}
    \rho(x, t) \, \dot{v}(x, t) = \nabla \cdot \tau(x, t) + f^{\text{ext}},
\end{equation}
where $\tau = \frac{1}{\det(F)} K (F^E) (F^E)^T$ with the Kirchhoff stress tensor $K=\frac{\partial \Psi}{\partial F}$ with a strain energy density $\Psi(F)$, and $K$ depends on the materials' elasticity models. The term $\nabla \cdot \tau(x, t)$ represents the internal forces within the material, while $f^{ext}$ denotes the external force.

To obtain the deformation map $\phi(X, t)$ while satisfying the mass and momentum conservation, we leverage the MPM \cite{Sulsky_1994}. This transforms the continuum into discrete Lagrangian particles carrying quantities such as velocity $v_i$, deformation gradient $F_i$, and position $x_i$. To achieve the two-way transfer of information between these Lagrangian particles and Eulerian grids, we utilize B-spline kernels with $C^1$ degree of continuity \cite{Sulsky_1994}. During the time step $t^n$ to $t^{n+1}$, the mass of Lagrangian particles remains constant under the mass conservation. With momentum conservation, the momentum of the particles also remains unchanged:
\begin{equation}
\frac{m_j}{\Delta t} (v_j^{n+1} - v_j^n) = - \sum_i V_i^0 \frac{\partial \Psi}{\partial F}(F^{E,n}_i) (F^{E,n}_i)^T \nabla \beta_{j,i}^n + f_j^{\text{ext}},
\end{equation}
where $i$ denotes the Lagrangian particles and $j$ represents the Eulerian grid; $m=\rho V$ is mass; $\beta$ is the B-spine kernel function; $V$ is volume. To update the Lagargian particles' positions, the updated Eulerian grid velocity $v^{n+1}_j$ is transferred onto $v^{n+1}_i$, then the particles' positions are updated as $x_{i}^{n+1} = x_{i}^{n} + \Delta t \, v_{i}^{n+1}$. The elastic deformation gradients of the particles are updated as $F^{E, n+1}_i = \left( I + \Delta t \, \sum_j \, v_{j}^{n+1} \, \nabla (\beta_{j,i}^n)^T \right) F^{E,n}_i$. Depending on the material-specific plasticity model, $F^{E, n+1}_i$ is adjusted by a mapping as $M: F^{E, n+1}_i \mapsto F^{E, n+1}_i$. Please refer to the supplementary document for detailed information on the plasticity mapping functions.

To apply the deformation map $\phi(X, t)$ to generate physics-grounded motion, we employ 3D Gaussians to represent the discrete particles. Under the assumption that particles undergo local affine transformations, which ensures the deformed Gaussian kernel remains Gaussian in the world space, the deformation map is approximated with the first-order Taylor expansion as $\tilde{\phi}_i(X, t) = x_i + F_i (X - X_i)$. The deformed Gaussian kernel then becomes:
\begin{equation}
    \begin{aligned}
    G_i(x, t) &= e^{-\frac{1}{2} (\tilde{\phi}^{-1}(x, t) - X_i)^T \Sigma_i^{-1}(\tilde{\phi}^{-1}(x, t) - X_i)} \\
    &= e^{-\frac{1}{2} (x - x_i)^T (F_i \Sigma_i F_i^T)^{-1} (x - x_i)}
    \end{aligned}
\end{equation}
Given the 3D Gaussians with $\{X_i, \Sigma_i, \alpha_i, c_i\}$, the deformation map $\phi(X, t)$ deforms them to $\{x_i(t), \sigma(t), \alpha_i, c_i\}$, where $x_i(t)=\phi(X_i, t)$ and $\sigma_i(t)=F_i(t) \Sigma_i F_i(t)^T$. 

To ensure consistent and accurate RGB color values during rendering, we regularize the RGB values converted from spherical harmonics $c_i$ through a color regularization process that includes the normalization and clamping steps. 

\noindent \textbf{Normalization.} The normalization step rescales the RGB values to fit within the [0, 1] range. This is achieved by subtracting the minimum value of the original RGB tensor and dividing by the difference between the maximum and minimum values, as follows:
\begin{equation}
    s_{\text{norm}} = \frac{s - s_{\text{min}}}{s_{\text{max}} - s_{\text{min}}},
\end{equation}
where $s$ represents the original RGB tensor, and $s_{min}$, $s_{max}$ are the minimum and maximum values in $s$, and $s_{norm}$ is the normalized RGB tensor.

\noindent \textbf{Clamping.} While the normalization step adjusts the RGB values to [0, 1], numerical precision issues can sometimes cause minor deviations, leading to values outside this range. For instance, these issues might arise when rounding errors accumulate to allow some values to exceed the [0, 1] range, and when normalization does not precisely yield values within [0, 1]. To address this, a clamping step is applied to enforce strict adherence to [0, 1]. The clamping operation is defined as:
\begin{equation}
    s_{\text{clamp}} = \min(\max(s_{\text{norm}}, 0), 1).
\end{equation}

With the RGB values regularized, a GS rasterizer is used to render the deformed Gaussian kernels. This process produces high-quality 3D objects with realistic physics-grounded motion.

\section{Experiments}
\label{sec:experiments}

In this section, we evaluate the effectiveness of our proposed framework through comprehensive experiments. We provide both qualitative and quantitative evaluations, along with our detailed ablation study results.

\subsection{Implementation Details}
\textbf{Setup.} We utilize Pytorch to implement the 3D Gaussian Splatting, adhering to the optimization pipeline from \cite{kerbl20233dgaussiansplattingrealtime}. To generate physics-grounded motion, we build upon the MPM \cite{Sulsky_1994, xie2024physgaussianphysicsintegrated3dgaussians}. Our experiments are conducted on an Nvidia RTX 3090 GPU.

\noindent \textbf{Metrics.} We utilize the LAION aesthetic score \cite{Huang_2024_CVPR}, which evaluates the aesthetic quality of a video on a scale from 0 to 10. Furthermore, we employ the CLIP score \cite{radford2021learningtransferablevisualmodels} to measure the prompt consistency of a video, which is the average cosine similarity between input prompt and all video frames. In addition, we adopt the Mean Opinion Score (MOS) for the human study evaluations on the generated videos.

\noindent \textbf{Method comparison.} As the first approach to utilize text for generating 3D objects with physics-grounded motion, there is no existing method for direct comparison. Due to the generative nature of our framework, the ground truth of the deformed scenes is also unavailable. To evaluate our method, we provide both qualitative and quantitative results of our framework and compare them against the results of the relevant 3D-to-motion methods \cite{huang2024dreamphysicslearningphysicalproperties, qiu2024featuresplattinglanguagedrivenphysicsbased}, using the 3D models provided by our approach.

\subsection{Qualitative Evaluation}
\begin{figure*}[htbp]
  \centering
  \includegraphics[width=\textwidth, trim=0 0 0 0, clip]{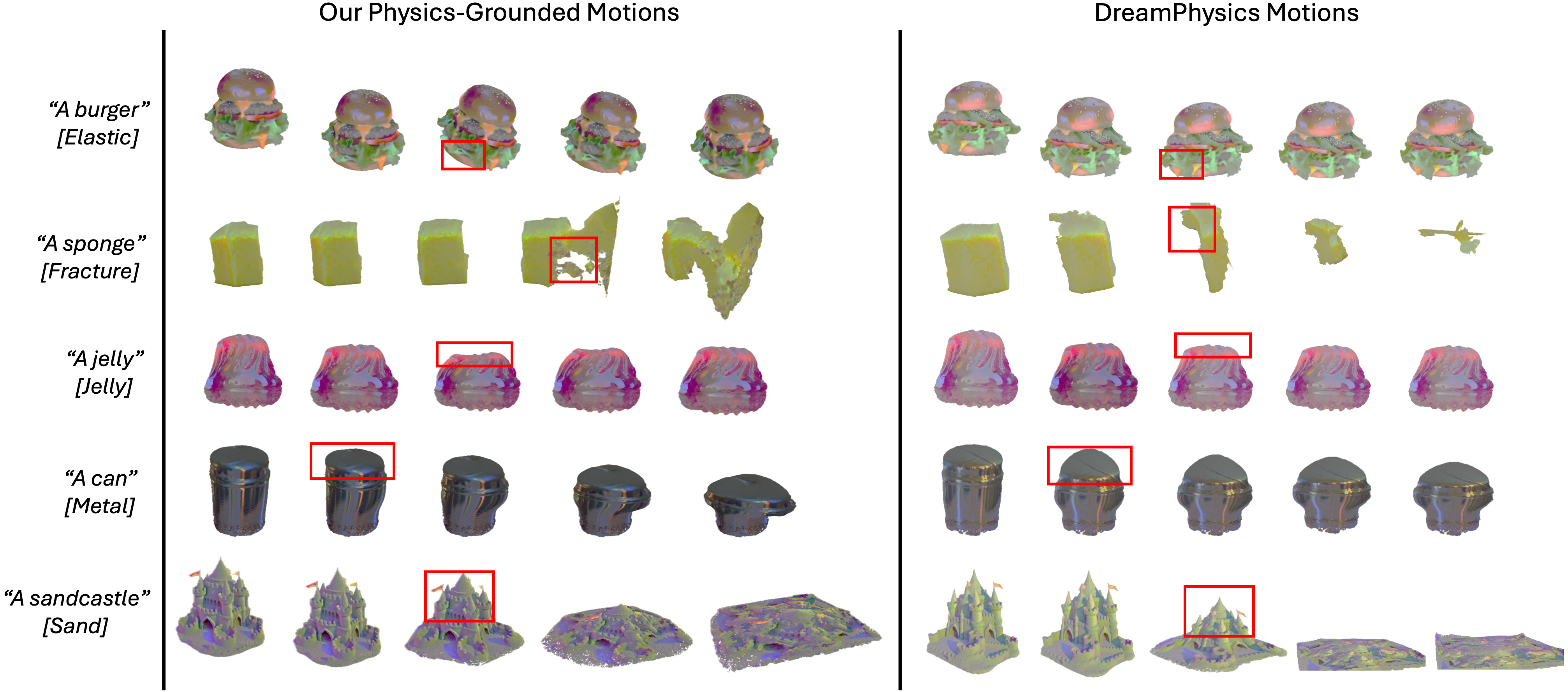}
  \captionsetup{belowskip=-10pt} 
  \caption{\textbf{Our results and DreamPhysics results.} We present our text-to-3D physics-grounded motion results and the results generated by DreamPhysics using the 3D models provided by our framework.}
  \label{qua_ours_dp}
\end{figure*}

\begin{figure}[htbp]
  \centering
  \includegraphics[width=\columnwidth]{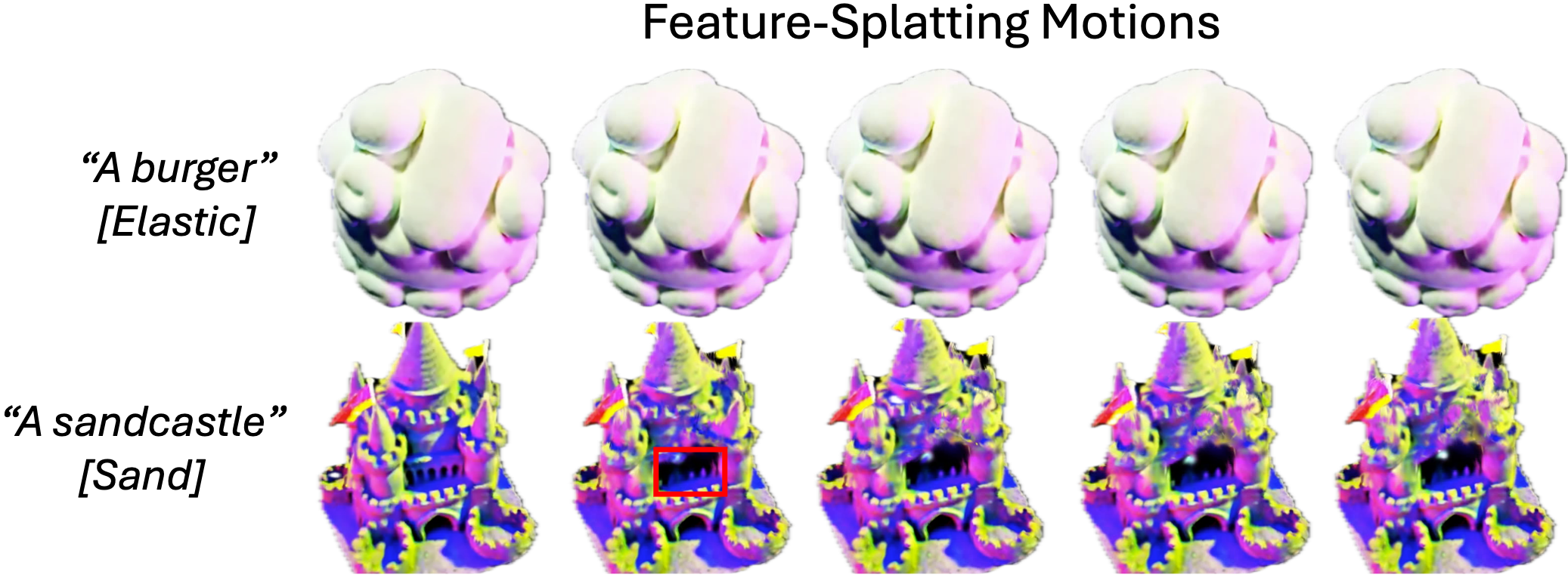}
  \captionsetup{belowskip=-10pt} 
  \caption{\textbf{Feature-Splatting results.} Results generated by Feature-Splatting using the 3D models provided by our framework.}
  \label{qua_fs}
\end{figure}

We showcase the qualitative performance of our framework by creating high-quality 3D objects featuring diverse physics-based dynamics. For each material dynamics type, we present an example illustrating the deformation motion of a 3D object, as shown in Figure~\ref{qua_ours_dp}. Please review the supplementary document for the material-specific elasticity and plasticity dynamics models. 

Our framework demonstrates the following dynamics: \textbf{elastic}, \textbf{fracture}, \textbf{jelly}, \textbf{metal}, and \textbf{sand}. Elastic and jelly dynamics refer to the behavior of objects where the rest shape remains unchanged despite undergoing deformation. Metal dynamics cause the object to undergo permanent deformation when its stress reaches a certain threshold, governed by von Mises plasticity model \cite{ROY2023106049}. Fracture dynamics simulate particle separation into clusters under significant deformation. Sand dynamics, modeled by Druker-Prager plasticity model \cite{10.1145/2897824.2925906}, captures granular level frictional effects among particles. As illustrated in Figure~\ref{qua_ours_dp}, our results demonstrate the capability of our framework to synthesize 3D objects with high-quality appearances, accurate shapes, and realistic physics-grounded motion for the given text prompts and material types.

\noindent \textbf{Qualiative comparisons.} Since our framework is the first to bridge text-to-3D synthesis with physics-grounded motion, there are no existing works available for direct qualitative comparison. Therefore, we evaluate the qualitative performance of our framework in comparison to the 3D-to-motion methods DreamPhysics \cite{huang2024dreamphysicslearningphysicalproperties} and Feature-Splatting \cite{qiu2024featuresplattinglanguagedrivenphysicsbased}, using our provided 3D models. Their corresponding results are shown in Figure~\ref{qua_ours_dp} and Figure~\ref{qua_fs}. Compared to the results of our framework, we observe that the results produced by DreamPhysics have under-saturated colors with weaker and less realistic motion. For instance, its fracture motion for the sponge object fails to achieve actual fracturing; instead, the object merely shrinks. Additionally, DreamPhysics’s generated metal motion for the can object lacks a realistic metallic movement when force is applied, particularly at the top of the can. Feature-Splatting, on the other hand, generates over-saturated and often inaccurate colors, with its motions being nearly imperceptible. In addition, it struggles with video segmentation in certain cases, such as the elastic burger example, which prevents successful motion generation. Moreover, it can only generate motion for elastic and sand material types, which has less versatility compared to our framework.

\begin{table}[htbp]
\centering
\captionsetup{belowskip=-10pt} 
\resizebox{\columnwidth}{!}{%
\begin{tabular}{@{}l|ccc@{}}
\toprule
Criteria  & Ours & DreamPhysics  & Feature-Splatting   \\ \midrule
Appearance    & \textbf{2.87} & 2.58 & 2.36   \\
Shape Accuracy       &  \textbf{3.22}  & 3.07 & 2.82   \\
Motion Quality &    \textbf{3.04}   & 2.67 & 2.61   \\
\bottomrule
\end{tabular}%
}
\caption{User study MOS results. Best results are in \textbf{bold}.}
\label{mos}
\end{table}

\noindent \textbf{User study.} A user study is conducted to evaluate the human-perceived quality of the synthesized 3D object motion videos. In the study, we recruit 20 participants to evaluate the videos. For each material type—elastic, fracture, jelly, metal, and sand—we generate corresponding 3D object motion videos using our framework and DreamPhysics. In addition, we employ Feature-Splatting to generate its elastic and sand video results. To measure participant evaluations of the generated videos, we adopt the Mean Opinion Score, with ratings ranging from 1 (Bad) to 5 (Excellent). Participants are instructed to rate each video based on 3 criteria: 1) visual appearance; 2) shape accuracy; 3) motion quality. 

We observe from the MOS results (Table~\ref{mos}) that: 1) Our framework outperforms the other methods in visual appearance, shape accuracy, and video quality. 2) The MOS results are aligned with the evaluation results presented in the paper.

\begin{table}[htbp]
\centering
\captionsetup{belowskip=-10pt}
\resizebox{\columnwidth}{!}{%
\begin{tabular}{@{}l|cccc@{}}
\toprule
Method & LAION ↑ & CLIP Score ↑ & Resolution ↑ & Generation Time (min) ↓ \\ \midrule
DreamPhysics \cite{huang2024dreamphysicslearningphysicalproperties} & 3.77 & 0.269 & 800x800 & 3.58   \\
Feature-Splatting \cite{qiu2024featuresplattinglanguagedrivenphysicsbased} &  1.98 & 0.268 & 512x512 & 6.57 \\
Ours &  \textbf{3.80} & \textbf{0.278} & \textbf{1958x1090} & \textbf{1.7} \\
\bottomrule
\end{tabular}%
}
\caption{Results of quantitative evaluation. Best results are in \textbf{bold}.}
\label{tab:table_quan}
\end{table}

\subsection{Quantitative Evaluation}

We conduct quantitative comparison with other 3D-to-motion methods \cite{huang2024dreamphysicslearningphysicalproperties, qiu2024featuresplattinglanguagedrivenphysicsbased} using the 3D models generated by our framework. As presented in Table~\ref{tab:table_quan}, our method surpasses other frameworks across multiple metrics, including the mean LAION score, CLIP score, video resolution, and generation time. These results highlight that our framework can achieve higher aesthetic quality, better prompt-video consistency, improved visual quality, and faster generation time. In contrast, despite using the 3D models generated by our framework, both DreamPhysics and Feature-Splatting exhibit worse performance across all metrics. This indicates the superiority and efficacy of our framework in producing high-quality text-to-3D motion videos.

\subsection{Ablation Studies}

Ablation studies are conducted on our proposed framework to demonstrate the necessity of \textbf{1) LLM-prompt refinement}, \textbf{2) 3D diffusion prior guidance}, \textbf{3) RGB color regularization}. The presented ablation results demonstrate the contribution of each of these components in improving the overall performance of our system. From Table~\ref{tab:abla_quan}, we observe that each proposed component plays an important role in ensuring both high aesthetic quality and strong prompt-video consistency. Overall, it shows that 3D diffusion prior guidance produces a larger impact on the LAION score, which indicates its importance in obtaining a high aesthetic quality. On the other hand, LLM-prompt refinement has a greater effect on the CLIP score, reflecting its ability to achieve semantically aligned generation results.

\begin{figure}[htbp]
  \centering
  \includegraphics[width=0.8\columnwidth]{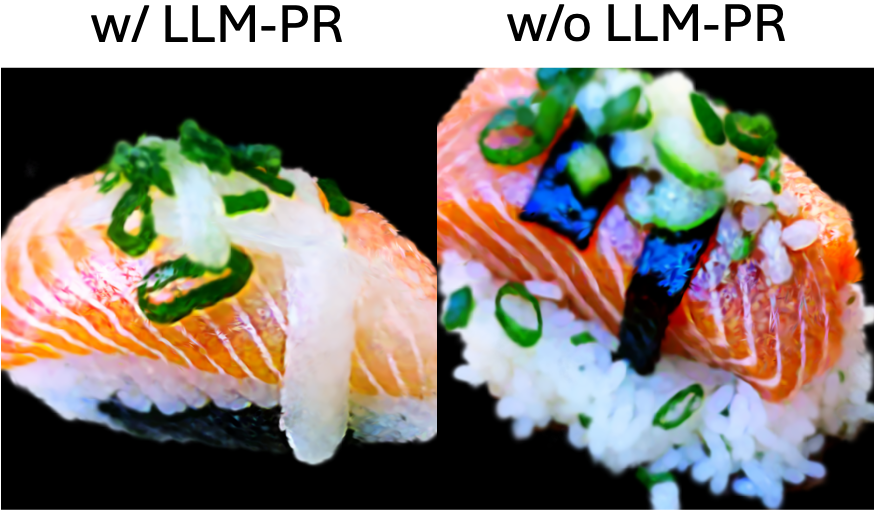}
  \captionsetup{belowskip=-15pt} 
  \caption{The impact of adopting LLM-prompt refinement. Prompt: \textit{A salmon nigiri.}}
  \label{abla_llm}
\end{figure}

\noindent \textbf{LLM-prompt refinement.} Figure~\ref{abla_llm} illustrates that without LLM-prompt refinement (LLM-PR), the generated object suffers from inaccuracies in object details. For instance, in the result generated without LLM-PR, the coloration of the salmon is inconsistent with its natural appearance, and the rice grains are inaccurately positioned on top of the salmon, reducing the overall realism and coherence of the scene. Quantitative ablation results in Table~\ref{tab:abla_quan} also shows that the absence of LLM-PR leads to reduced LAION score and CLIP score, which reflects the critical role of LLM-PR in achieving a high aesthetic quality and prompt-video consistency.
 
\begin{figure}[htbp]
  \centering
  \includegraphics[width=\columnwidth]{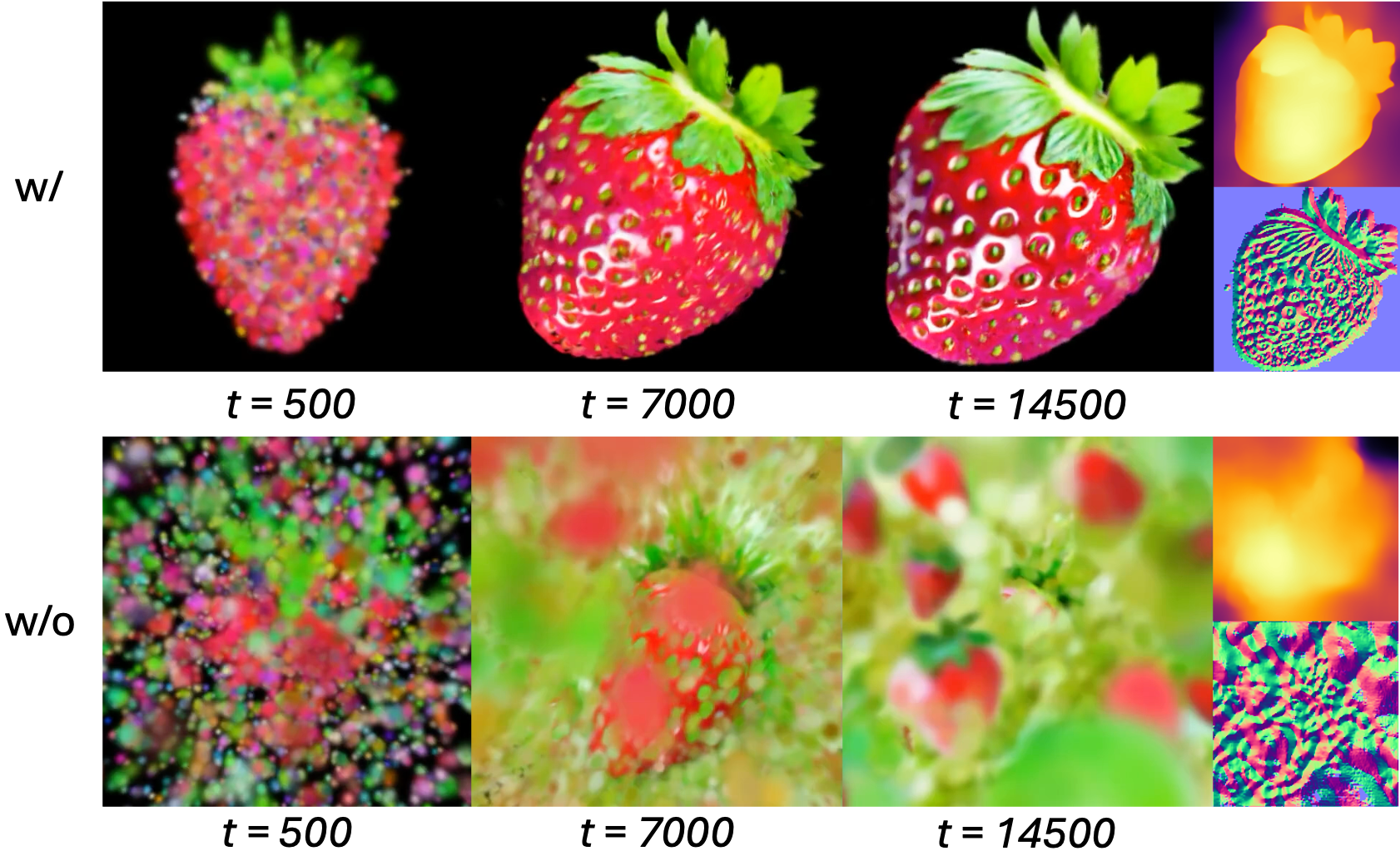}
  \captionsetup{belowskip=-15pt} 
  \caption{The impact of employing the 3D diffusion prior as GS shape guidance. Prompt: \textit{A strawberry.}}
  \label{abla_guide}
\end{figure}

\noindent \textbf{3D diffusion pior guidance.} The results in Figure~\ref{abla_guide} demonstrate that the incorporation of the 3D diffusion prior as guidance (3D Guidance) enables the Gaussian Splatting to generate the 3D object with a more accurate geometrical shape, compared to the results generated without this guidance. Furthermore, as presented in Table~\ref{tab:abla_quan}, the absence of the 3D diffusion prior guidance results in lower LAION and CLIP scores. This indicates the importance of 3D Guidance in improving both the aesthetic quality and the coherence between the input prompt and the resulting video.

\begin{figure}[htbp]
  \centering
  \includegraphics[width=0.65\columnwidth]{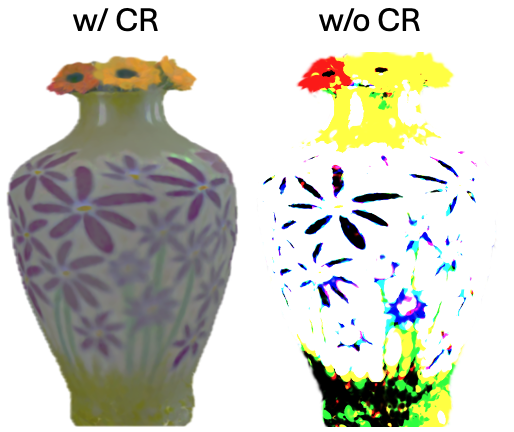}
  \captionsetup{belowskip=-15pt} 
  \caption{The effects of applying the color regularization on the RGB values. Prompt: \textit{A vase with flowers.}}
  \label{abla_color}
\end{figure}

\noindent \textbf{Color regularization.} Our results in Figure~\ref{abla_color} indicate that the absence of the color regularization (CR) leads to the inaccurately rendered colors of the 3D object, compared to the results generated with the color regularization. This is also reflected in Table~\ref{tab:abla_quan}, because the LAION score and CLIP score in the results generated without CR are noticeably reduced. This shows that CR plays a significant role in obtaining accurate color rendering, which also ensures a high aesthetic quality and prompt consistency of the generated video.

\begin{table}[htbp]
\centering
\captionsetup{belowskip=-10pt}
\resizebox{\columnwidth}{!}{%
\begin{tabular}{@{}l|cc@{}}
\toprule
Method & LAION ↑ & CLIP Score ↑ \\ \midrule
w/o LLM-PR & 3.62 & 0.256 \\
w/o 3D Guidance & 3.29 & 0.263 \\
w/o CR & 3.48 & 0.266 \\ \midrule
Full & \textbf{3.80} & \textbf{0.278} \\
\bottomrule
\end{tabular}%
}
\caption{Ablation study results. Best results are in \textbf{bold}.}
\label{tab:abla_quan}
\end{table}

\section{Discussion}
\label{sec:discussion}

\noindent \textbf{Limitation.} Our framework does not currently support rendering the interaction of 3D object surfaces with light, so it cannot generate the effects such as reflections or shadows. Additionally, our framework only supports the motion simulation of limited material types. Future work could explore integrating advanced relighting techniques and expanding the range of material types to enhance the framework's versatility and realism.

\noindent \textbf{Conclusion.} 
In this paper, we present an innovative framework for text-to-3D motion generation based on physics, facilitating the creation of high-quality 3D objects with realistic, physics-aware movements, effectively integrating generative modeling with physics-driven motion simulation. Our framework integrates four innovative components: 1) \textbf{LLM-prompt refinement} to ensure the accurate 3D generation for the prompt; 2) \textbf{diffusion prior guidance} for steering the generative process toward the result with accurate shape and high-quality visual appearance; 3) \textbf{continuum mechanics-based deformation mapping} to  to model realistic physical interactions and deformations of the generated 3D object; 4) \textbf{color regularization} for consistent and accurate color rendering. This unified pipeline integrates natural language processing, generative modeling, and physics simulation to redefine the boundaries of 3D content creation, which paves the way for transformative applications across diverse industries such as filmmaking, virtual/augmented reality, gaming, and beyond.

\clearpage   
{
    \small
    \bibliographystyle{ieeenat_fullname}
    \bibliography{main}
}

\clearpage
\setcounter{page}{1}
\maketitlesupplementary

\section{Theoretical Details}
\label{sec:theoretical_details}

\subsection{The Material Point Method}
The MPM is a computational physics method to simulate the material behaviors under different physical forces and deformations \cite{Sulsky_1994}. The MPM discretizes a material body into a collection of Lagrangian particles, and each particle has a set of quantities such as position $x^n_i$, mass $m_i$, velocity $v^n_i$, Kirchhoff stress tensor $K^n_i$, deformation gradient $F^n_i$, and affine momentum $A^n_i$ on particle $i$ at time $t^n$. At time $t^n$, let $x^n_j$, $m_j$, and $v^n_j$ represent the position, mass, and velocity on grid node $j$. These grid nodes facilitate the calculation of the deformations and the applied forces on the material body. Due to the conservation of mass, particle mass is invariant. At each time step, the MPM conducts a two-way transfer: 1) Particle-to-Grid; 2) Grid-to-Particle.

\noindent \textbf{Particle-to-Grid Transfer.} In this process, the mass and particle momentum are transferred to the grids \cite{xie2024physgaussianphysicsintegrated3dgaussians}. The mass $m^n_j$ at a grid node $j$ is computed as:
\begin{equation}
    m^n_j = \sum_i w^n_{ji} m_i,
\end{equation}
where $w^n_{ji}$ is the interpolation weight obtained from a B-spline kernel. Using the APIC momentum transfer method \cite{10.1145/2766996}, the momentum at the grid node $j$ is updated as:
\begin{equation}
    m^n_j v^n_j = \sum_i w^n_{ji} m_i (v^n_i + A^n_i (x_j -x^n_i)).
\end{equation}
Based on the particles' internal and external forces, the grid velocity $v^{n+1}_j$ at the next time step is updated as:
\begin{equation}
    v_{j}^{n+1} = v_{j}^{n} - \frac{\Delta t}{m_j} \sum_{i} K_{i}^{n} \nabla w_{ji}^{n} V_{i}^{0} + \Delta t  g,
\end{equation}
where $g$ is the gravity acceleration.

\noindent \textbf{Grid-to-Particle Transfer.} In this stage, the grid nodes' updated velocities and momentum are transferred back to the particles \cite{xie2024physgaussianphysicsintegrated3dgaussians, 10.1145/2897826.2927348}. The velocity $v^{n+1}_i$, position $x^{n+1}_i$, affine momentum $A^{n+1}_i$, and deformation gradient $F^{n+1}_i$ of particle $i$ at the new time step are updated as:
\begin{equation}
\begin{aligned}
&v^{n+1}_i = \sum_j v^{n+1}_j w^n_{ji}, \\
&x_{i}^{n+1} = x_{i}^{n} + \Delta t \, v_{i}^{n+1},\\
&A^{n+1}_i = \frac{12}{\Delta x^2 (b + 1)} \sum_{j} w_{ji}^{n} v_{j}^{n+1} \left( x_{j}^{n} - x_{i}^{n} \right)^{T},\\
&\nabla v_{i}^{n+1} = \sum_{j} v_{j}^{n+1} \left( \nabla w_{ji}^{n} \right)^{T},\\
&F_{i}^{n+1} = M(\left( I + \nabla v_{i}^{n+1} \right) F_{i}^{n}).
\end{aligned}
\end{equation}
Here, $b$ denotes the B-spling degree, and $\Delta x$ represents the Eulerian grid spacing. The calculation of the deformation adjustment mapping $M$ and the Kirchhoff stress tensor $K$ are detailed in the next subsection.
\begin{table}[t]
  \centering
  \captionsetup{belowskip=-15pt} 
  \resizebox{\columnwidth}{!}{%
  \begin{tabular}{@{}l|c|c@{}}
    \toprule
    Notation & Meaning & Definition \\
    \midrule
    $E$ & Young's modulus & - \\
    $\mu$ & Shear modulus & $\mu = \frac{E}{2(1+\nu)}$ \\
    $\lambda$ & Lam\'{e} modulus & $\lambda = \frac{E \nu}{(1 + \nu)(1 - 2\nu)}$ \\
    \bottomrule
  \end{tabular}
  }
  \caption{Material Parameters.}
  \label{tab:mat_parameters}
\end{table}

\subsection{Physics Models}

In this section, we provide the physics model details for the paper, and we show the relevant material parameters in Table~\ref{tab:mat_parameters}. The employed physics models are adopted from \cite{10.1145/3610548.3618207, xie2024physgaussianphysicsintegrated3dgaussians}. For the given material, the Kirchhoff stress tensor $K$ is mapped by the material's corresponding elasticity model, and the deformation gradient $F^E$ is mapped by the material's specific plasticity model.

\noindent \textbf{Fixed Corotated Elasticity.} The Fixed Corotated Elasticity model describes the behaviors of materials that undergo deformations with rotations and small elastic strains \cite{10.1145/2766996}:
\begin{equation}
K = 2\mu \left(F^E - R\right) (F^E)^T + \lambda \left(J - 1\right) J,
\end{equation}
where $R=UV^T$ and $F^E=U\Sigma V^T$, and $J$ is the determinant of $F^E$.

\noindent \textbf{St. Venant-Kirchhoff Elasticity.} St. Venant-Kirchhoff models materials that return to their original shapes after large deformations \cite{10.1145/2897824.2925906}:
\begin{equation}
    K = U\left(2 \mu \epsilon + \lambda \text{sum}(\epsilon)1\right) V^T,
\end{equation}
where $\epsilon=\log(\Sigma)$ and $F^E=U \Sigma V^T$.

\noindent \textbf{Drucker-Prager Plasticity.} Drucker-Prager Plasticity describes the behaviors of the materials that do not exhibit purely ductile behavior \cite{10.1145/2897824.2925906}:
\begin{equation}
    F^E = U M(\Sigma) V^T,
\end{equation}
\begin{equation}
    M(\Sigma) = 
\begin{cases} 
    1 & \text{sum}(\epsilon) > 0 \\ 
    \Sigma & \delta_\gamma \leq 0 \text{ and } \text{sum}(\epsilon) \leq 0 \\ 
    \text{exp}(\epsilon - \delta \gamma \frac{\hat{\epsilon}}{\|\hat{\epsilon}\|}) & \text{otherwise}.
\end{cases}
\end{equation}
Here, $M$ is the deformation adjustment mapping, $\delta \gamma=\|\hat{\varepsilon}\| + \alpha \frac{(d\lambda + 2\mu)\text{sum}(\epsilon)}{2\mu}$, $\alpha = \sqrt{\frac{2}{3}} \frac{2\text{sin} \phi_f}{3-\text{sin}\phi_f}$, $\phi_f$ is the friction angle, and $\hat{\epsilon} = \text{dev}(\epsilon)$.

\noindent \textbf{von Mises Plasticity.} von Mises Plasticity models the materials that will permanently deform when the stress reaches a certain threshold value \cite{ROY2023106049}:
\begin{equation}
    F^E = U M(\Sigma) V^T,
\end{equation}
\begin{equation}
    M(\Sigma) = 
\begin{cases} 
    \Sigma & \delta \gamma \leq 0 \\ 
    \text{exp}(\epsilon - \delta \gamma \frac{\hat{\epsilon}}{\|\hat{\epsilon}\|}) & \text{otherwise},
\end{cases}
\end{equation}
where $\delta \gamma = \| \hat{\epsilon} \|_F - \frac{KY}{2\mu}$, and $KY$ is the yield stress.

\subsection{Score Distillation Sampling}

Score Distillation Sampling is a technique proposed in DreamFusion \cite{poole2022dreamfusiontextto3dusing2d} that utilizes the 2D diffusion prior to optimize an image generator based on the probability density distillation. To achieve this, an image generator parameterized by parameters $\theta$ is represented as $g(\theta)$. To optimize over parameters $\theta$ such that the generated image $x = g(\theta)$ resembles a sampling from the pre-trained frozen 2D diffusion model, the SDS loss gradient for optimizing $\theta$ is formulated as:
\begin{equation}
    \nabla_{\theta} L_{\text{SDS}}(\phi, x = g(\theta)) \overset{\mathrm{\Delta}}{=} \mathbb{E}_{t, \epsilon} \left[ w(t) \left( \hat{\epsilon}_{\phi}(z_t; y, t) - \epsilon \right) \frac{\partial x}{\partial \theta} \right],
\end{equation}
where $\hat{\epsilon}_{\phi}(z_t; y, t)$ is the predicted noise by the pre-trained 2D diffusion model with the text prompt $y$ at the time step $t$, and $\epsilon$ is the true noise at the time step. $\frac{\partial x}{\partial \theta}$ is the derivative of the image generator's generated image with respect to its parameters $\theta$, and $w(t)$ is a weighting function from DDPM \cite{ho2020denoisingdiffusionprobabilisticmodels}. This loss function aligns the scores (or gradients) of the image generator and the 2D diffusion model by optimizing the loss gradients with respect to $\theta$, which can enable the use of the 2D diffusion prior to guide the generation of 3D models efficiently.

\section{More Results}
\label{sec:additional_results}

\begin{figure*}[htbp]
  \centering
  \includegraphics[width=\textwidth, trim=0 0 0 0, clip]{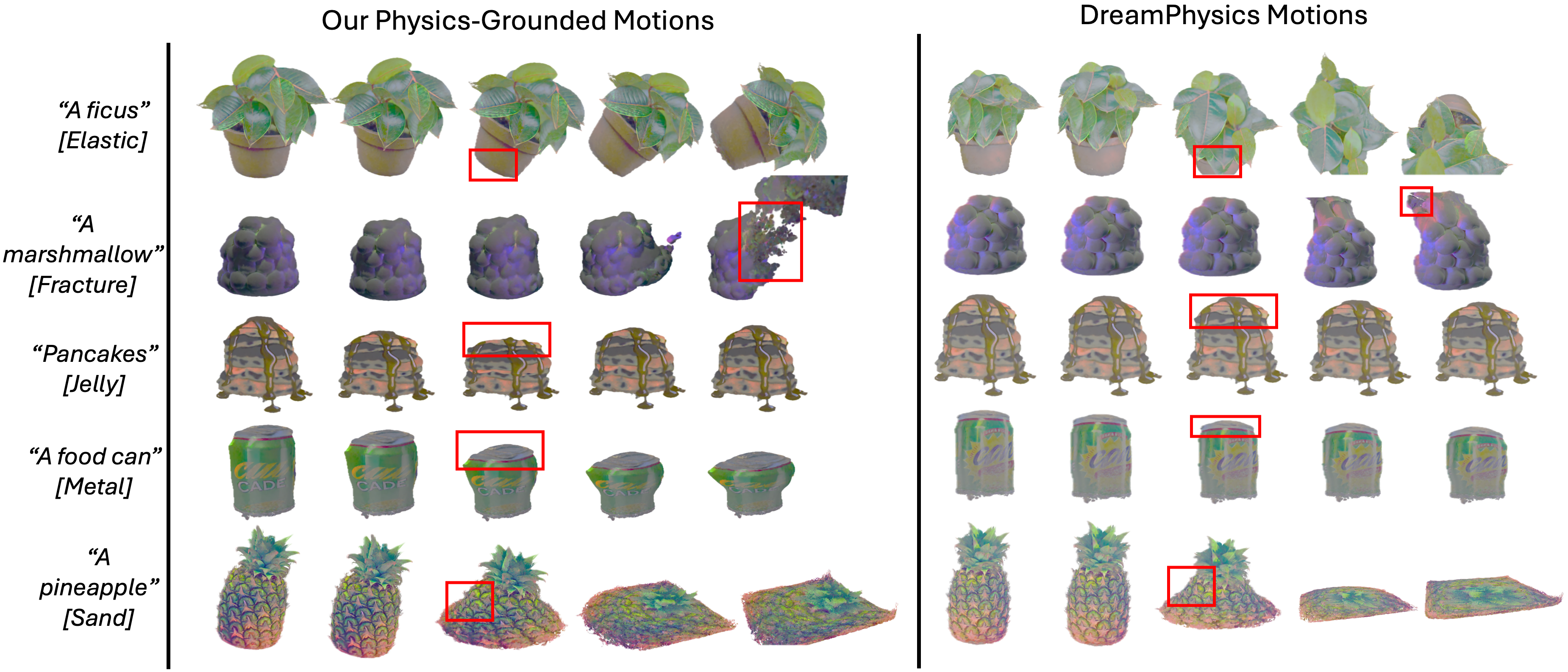}
  \captionsetup{belowskip=-10pt} 
  \caption{Additional text-to-3D physics-grounded motion results generated by our framework and the results generated by DreamPhysics using the 3D models provided by our framework.}
  \label{suppl_ours_dp}
\end{figure*}

\begin{figure}[htbp]
  \centering
  \includegraphics[width=\columnwidth]{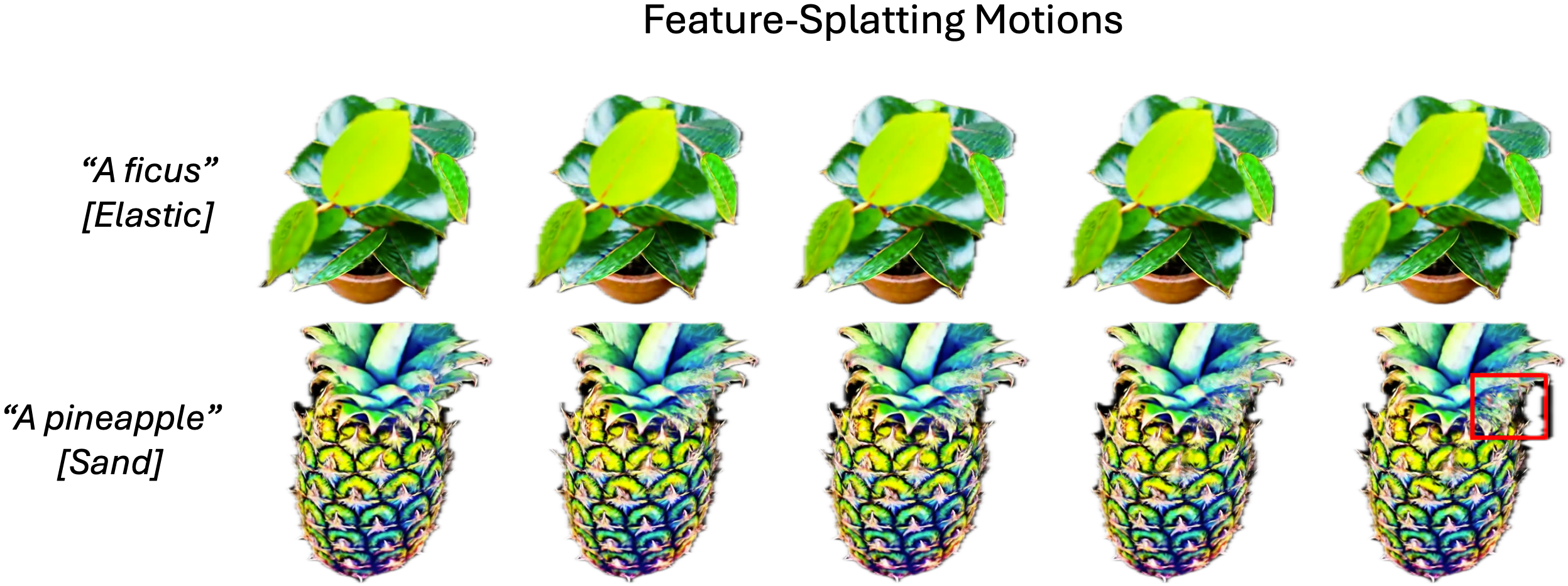}
  \captionsetup{belowskip=-10pt} 
  \caption{\textbf{Feature-Splatting Results.} Results generated by Feature-Splatting using the 3D models provided by our framework.}
  \label{suppl_fs}
\end{figure}

\begin{figure}[htbp]
  \centering
  \includegraphics[width=0.8\columnwidth]{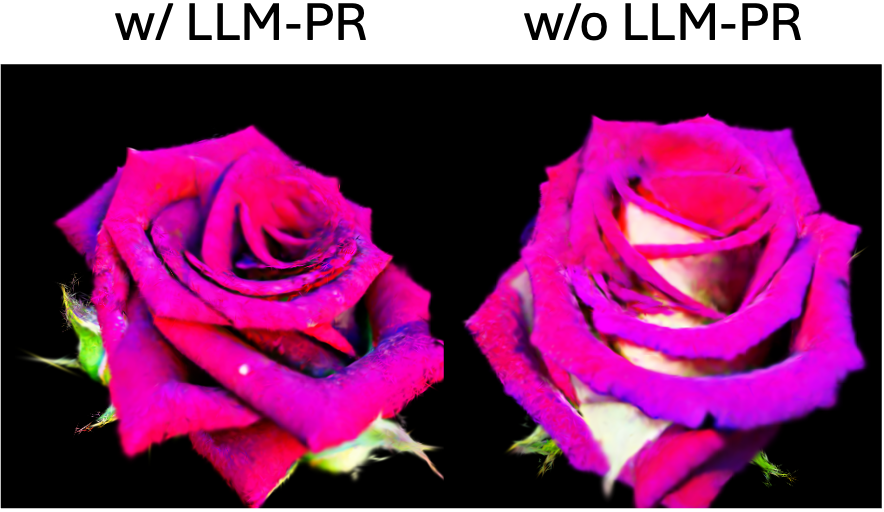}
  \captionsetup{belowskip=-10pt} 
  \caption{The impact of adopting LLM-prompt refinement. Prompt: \textit{A rose}.}
  \label{suppl_abla_llm}
\end{figure}

\begin{figure}[htbp]
  \centering
  \includegraphics[width=\columnwidth]{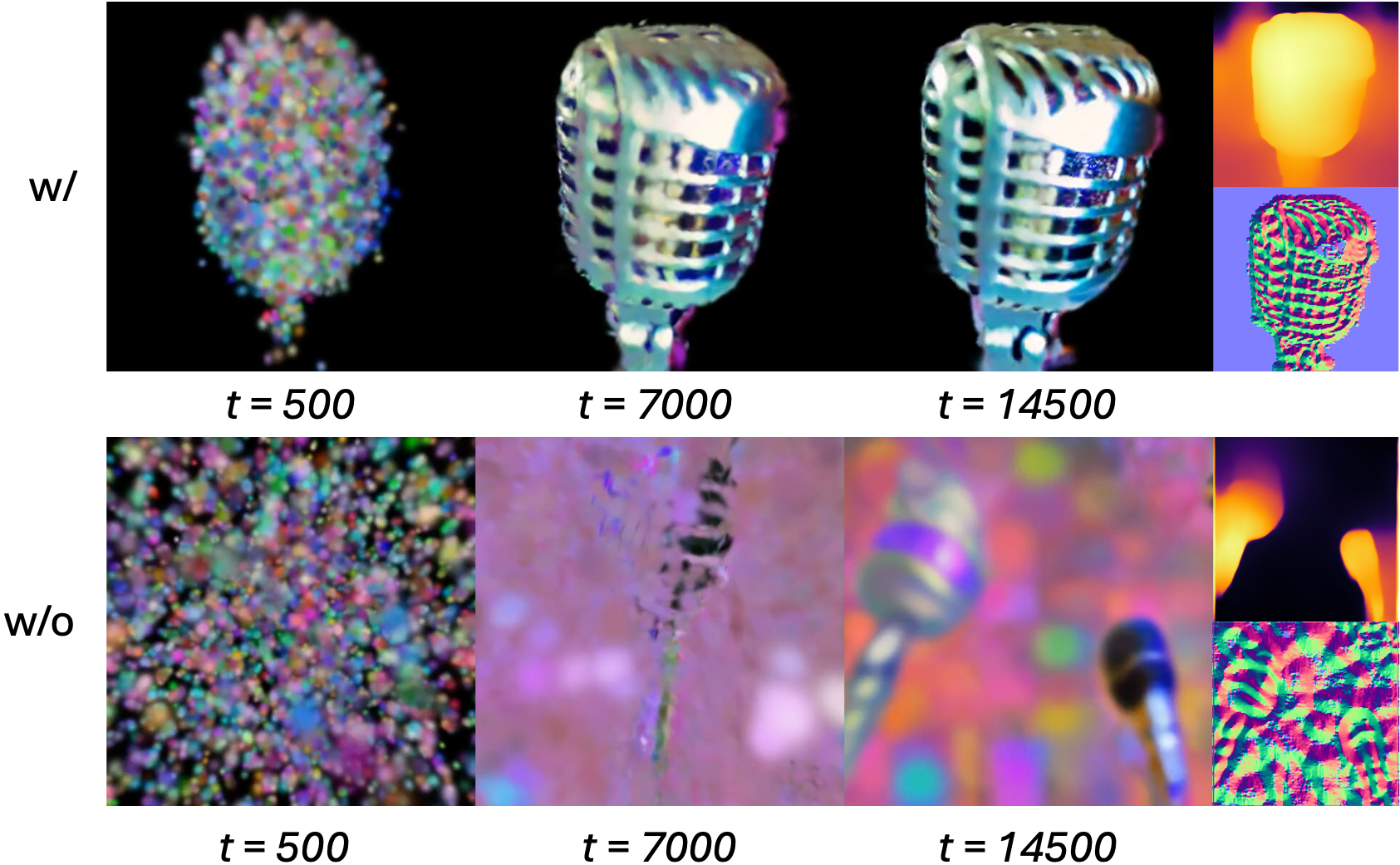}
  \captionsetup{belowskip=-10pt} 
  \caption{The impact of employing the 3D diffusion prior as GS shape guidance. Prompt: \textit{A microphone}.}
  \label{suppl_abla_3d}
\end{figure}

\begin{figure}[htbp]
  \centering
  \includegraphics[width=0.8\columnwidth]{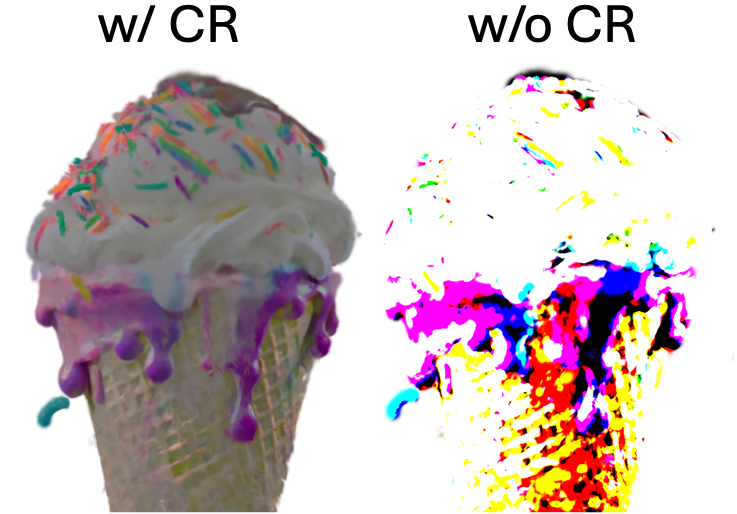}
  \captionsetup{belowskip=-10pt} 
  \caption{The effects of applying the color regularization on the RGB values. Prompt: \textit{An ice-cream}.}
  \label{suppl_abla_cr}
\end{figure}

\subsection{More Text-to-3D Physics Motion Results}
We present additional text-to-3D physics-grounded motion results in Figure~\ref{suppl_ours_dp}. These results demonstrate that our framework effectively generates 3D objects with high-quality appearances, accurate shapes, and realistic physics-driven motion for the given text prompts and material types. We also include the generated videos in the supplementary materials, please refer to the MP4 files in the videos folder or view them through the videos.html file.

\subsection{More Qualitative Comparisons}
The additional results of the compared methods, DreamPhysics \cite{huang2024dreamphysicslearningphysicalproperties} and Feature-Splatting \cite{qiu2024featuresplattinglanguagedrivenphysicsbased}, are presented in Figure~\ref{suppl_ours_dp} and Figure~\ref{suppl_fs}, respectively. Compared to our findings, we notice that DreamPhysics generates under-saturated colors, displaying muted and less convincing movements. For example, the jelly-like motion it produces for the pancakes is limited to minimal movements. Additionally, DreamPhysics's generated metal-like motion for the can model seems unnatural and less authentic. In contrast, Feature-Splatting produces over-saturated and inaccurate colors, with its generated motion being nearly undetectable.

\begin{table}
  \centering
  \captionsetup{belowskip=-10pt} 
  \resizebox{\columnwidth}{!}{
  \begin{tabular}{@{}c|cc@{}}
    \toprule
    Ours & DreamPhysics & Feature-Splatting \\
    \midrule
    \textbf{3.88} & 3.71 & 2.23 \\
    \bottomrule
  \end{tabular}
  }
  \caption{Mean LAION aesthetic scores of the 5 object videos generated by all methods.}
  \label{tab:table_laion_all_sup}
\end{table}
\subsection{More Quantitative Comparisons}
The additional quantitative comparison results are shown in Table~\ref{tab:table_laion_all_sup} for the object videos in Figure~\ref{suppl_ours_dp}. The table shows that our framework achieves a mean LAION score of 3.88, outperforming the other methods. This indicates that our framework produces videos with higher visual quality compared to the other methods.

\section{More Experiments}

\subsection{LLM-Prompt Refinement}
Figure~\ref{suppl_abla_llm} demonstrates that the absence of LLM-Prompt Refinement (LLM-PR) can lead to uneven and over-saturated colors, as well as the lack of shadows, fine textures, and intricate details in the flower petals.

\subsection{3D Diffusion Prior Guidance}
The results in Figure~\ref{suppl_abla_3d} illustrate that incorporating the 3D diffusion prior as guidance (3D Guidance) significantly improves the Gaussian Splatting process, enabling it to produce 3D objects with more precise geometrical shapes compared to those generated without this guidance.

\subsection{Color Regularization}
Our results in Figure~\ref{suppl_abla_cr} indicate that the absence of the color regularization (CR) on RGB values results in inaccurately rendered colors in the generated 3D object, as opposed to the more accurate results achieved with color regularization.

\end{document}